\documentclass[pdflatex,sn-mathphys]{sn-jnl}
\jyear{2022}

\usepackage{listings}
\usepackage[frozencache=true,cachedir=minted-cache]{minted}
\usepackage{caption}
\usepackage{amsmath}
\definecolor{codegreen}{rgb}{0,0.6,0}
\definecolor{codegray}{rgb}{0.5,0.5,0.5}
\definecolor{codepurple}{rgb}{0.58,0,0.82}
\definecolor{backcolour}{rgb}{0.95,0.95,0.92}

\lstdefinestyle{mystyle}{
    backgroundcolor=\color{backcolour},   
    commentstyle=\color{codegreen},
    keywordstyle=\color{magenta},
    numberstyle=\tiny\color{codegray},
    stringstyle=\color{codepurple},
    basicstyle=\ttfamily\footnotesize,
    breakatwhitespace=false,         
    breaklines=true,                 
    captionpos=b,                    
    keepspaces=true,                 
    numbers=left,                    
    numbersep=5pt,                  
    showspaces=false,                
    showstringspaces=false,
    showtabs=false,                  
    tabsize=2
}
\begin{document}

\title[Language Models are not Models of Language]{A Pr\'ecis of Language Models are not Models of Language}


\author*[]{\fnm{Csaba} \sur{Veres}}\email{csaba.veres@uib.no}



\affil[]{\orgdiv{Department of Information Science and Media Studies}, \orgname{University of Bergen}, \orgaddress{\city{Bergen}, \country{Norway}}}








\maketitle


\textbf{Natural Language Processing is one of the leading application areas in the current resurgence of Artificial Intelligence, spearheaded by Artificial Neural Networks. We show that despite their many successes at performing linguistic tasks, Large Neural Language Models are ill suited as comprehensive models of natural language. The wider implication is that, in spite of the often overbearing optimism about "AI", modern neural models do not represent a revolution in our understanding of cognition.}\newline

High level programming languages for digital computers, and theories of natural language have a curious historical connection. John W. Backus who led the Applied Science Division of IBM's Programming Research Group\footnote{\url{https://betanews.com/2007/03/20/john-w-backus-1924-2007/}} took inspiration from Noam Chomsky's work on phrase structure grammars and conceived a \emph{meta-language} that could specify the syntax of computer languages that were easier for programmers to write than assembler languages. The meta language later became known as \emph{Backus-Naur} form (BNF), so called partly because it was originally co-developed by Peter Naur in a 1963 IBM report on the ALGOL 60 programming language"\footnote{\url{https://www.masswerk.at/algol60/report.htm}}. The BNF is a notation for context free grammars consisting of \emph{productions} over \emph{terminal} and \emph{nonterminal} symbols, which defines the grammar of programming languages required for writing compilers and interpreters \cite{aho}.

Natural language is of course different from programming languages in many ways, not the least of which is that the grammar of programming languages is perfectly known, whereas the role of generative grammar in natural language is merely a hypothesis. Chomsky characterised Language as a set of sentences (potentially infinite) constructed out of a finite set of elements following the rules of a grammar. The role of Linguistics as a science, then, is to discover grammars that are able to distinguish legal productions which are part of the Language from ill formed strings that are not \cite{structures}. When a string of words is deemed unacceptable by a native speaker then this is the result, by hypothesis, of a violation of grammatical constraints. Similarly, the set of written statements in programming languages are productions of the grammar defined for the language. When a programmer writes code which does not compile or execute, then it is likely that they have violated the grammar and the compiler is unable to parse the code \cite{aho}. 

The claim that grammar has a central role in Natural Language has been questioned as a result of the success of \emph{Transformer} based neural \emph{Language Models} (LMs) \cite{vaswani2017attention}, which have acquired significant competence in various natural language tasks, including judgement of grammatical acceptability \cite{warstadt2018neural}.

Neural LMs are based on traditional statistical n-gram language models which are joint probability distributions over sequences of words, or alternatively, functions that return a probability measure over strings drawn from some vocabulary \cite{manning2008introduction}. More informally, language models can refer to "any system trained only on the task of string prediction" \cite{bender2020/v1/2020.acl-main.463} (p. 5185). Large neural LMs advance n-gram models by learning probability functions for sequences of real valued, continuous vector representations of words rather than the discrete words themselves. Continuous representations are effective at generalising across novel contexts, resulting in better performance across a range of tasks \cite{bengio2003}. Manning \cite{manning/coli_a_00239} describes several ways in which Deep Learning models can challenge traditional grammar based approaches in the theoretical understanding of Language.

Bengio et. al. \cite{hinton/turing} went further in arguing that continuous representations in Deep Learning models fundamentally differentiate neural LMs from traditional symbolic systems such as grammar because they enable computations based on non-linear transformations between the representing vectors themselves. As an example, "If Tuesday and Thursday are represented by very similar vectors, they will have very similar causal effects on other vectors of neural activity." \cite{hinton/turing} (p.59). In a Classical symbolic system there is no inherent similarity between the two symbols "Tuesday" and "Thursday", and "similar causal effects" must be prescribed by explicit axioms (see \cite{fodor88/0010-0277(88)90031-5} for a deep dicussion on the fundamental differences between symbolic and distributed architectures.). Large neural LMs are therefore a fundamental challenge to rule based theories because they obviate the need for explicit rules. 

Pinker and Prince \cite{pinker/prince} designated neural approaches which eschew traditional rules as \emph{eliminative connectionism}. In eliminative (neural) systems it is impossible to find a principled mapping between the components of the distributed (vector) processing model and the steps involved in a symbol-processing theory. Note that neural systems are not \emph{necessarily} eliminative. \emph{Implementational connectionism} is a class of systems where the computations carried out by collections of neurons are isomorphic to the structures and symbol manipulations of a symbolic system. For example, \emph{recurrent neural networks} with long short-term memory have been shown to learn very simple context free and context sensitive languages. Thus, the language with sentences of the form $a\textsuperscript{n}b\textsuperscript{n}$ can be learned with \emph{gate units} acting as counters that can keep track of the number of terminal strings in simple sequences \cite{gers}. Crucially, an implementational system could be fully compatible with a symbol based grammatical theory, and a network architecture that can induce grammatical principles would have minimal impact on our understanding of language. Pinker and Prince argued that language is a "crucial test case" for eliminative connectionism because so much of our understanding of language is bound up in the symbolic paradigm. In this commentary we argue that neural models of programming languages can provide an even more crucial test since we \emph{know} that computer code is governed completely by the symbolic paradigm.

Deep Learning neural networks have been shown to generate computer code. For example OpenAI Codex\footnote{\url{https://openai.com/blog/openai-codex/} }, an experimental API which powers GitHub Copilot\footnote{\url{https://copilot.github.com/}} and based on the GPT-3 language model fine tuned on publicly available computer code, can generate Python code from short textual docstrings \cite{chen2021codex}. Listing 1 shows a simple example problem from the evaluation set (including function signature, docstring, body, and several unit tests) together with a correct and an incorrect solution for the (simple) function \textbf{is\textunderscore prime}. 

\begin{lstlisting}[language=Python, caption=Example problem specification with a correct and an incorrect completion]    
def is_prime(n): 
    """Return true if a given number is prime, and 
        false otherwise. 
    >>> is_prime(6) False 
    >>> is_prime(101) True 
    >>> is_prime(11) True 
    >>> is_prime(13441) True 
    >>> is_prime(61) True 
    >>> is_prime(4) False 
    >>> is_prime(1) False 
    """
    ### COMPLETION 6 (CORRECT): ###
    prime = True
    if n == 1:
        return False
    for i in range(2, n):
        if n % i == 0:
            prime = False
    return prime
    ### COMPLETION 7 (WRONG): ###
    for i in range(2, n):
        if n % i == 0:
            return False
    return True
    
\end{lstlisting}

Codex generated functionally correct code on up to 60\% of the problems. However, the solutions were syntactically correct in over 90\% of cases, showing that errors were predominantly semantic \cite{chen2021codex}. Often the proposed solution solved only a part of the problem as in this example, where the incorrect algorithm fails to consider the boundary condition where n = 1. 

Austin et al. \cite{austin} constructed a slightly more difficult dataset, the Mostly Basic Programming Problems (MBPP) which were used to test BERT-style transformer models \cite{vaswani2017attention} with parameter counts ranging from 244 million to 137 billion. The smallest models produced syntactically correct Python code approximately 80\% of the time, increasing to over 90\% for the larger models. 

LMs wich produce computer code bring into sharp focus the nature of explanation in neural models. In order to generate code, one possibility is that networks learn the grammar of the language(s) they are exposed to. There is some support for this in evidence of syntactic information in natural language word representations \cite{hewitt-manning-2019-structural}. However this evidence is far short of an argument that language \emph{rules} are learned. More importantly, even if this were eventually shown to be the case, the conclusion would be that LMs are \emph{implementational} after all, and their theoretical interest would focus on their ability to learn rules without explicit instruction. Such models can not provide more insight into the natural phenomena than we already have. In the case of computer languages they provide no principled reason for why some strings are syntactically valid and some are not. In reality this is determined entirely by the grammar. 

The second possibility is that LMs are simply learning sophisticated statistical properties of their training data and extrapolate based on the learned model \cite{extrapolate}. On this view the success of LM architectures in generating computer code shows just how well they are able to extrapolate, being able to mimic the productions of a formal system without knowledge of its rules. In the absence of arguments to the contrary there is no reason to think that their performance on natural language tasks is any different. That is, large language models are simply extrapolating from their training data and have nothing to say about the claim that natural language is governed by a grammar.

Pinker and Prince argued that the connectionist models of the time failed to deliver a "radical restructuring of cognitive theory" (\cite{pinker/prince}, p.78) because they did not adequately model the relevant linguistic phenomena. We argue that modern neural models similarly fail, but from the opposite perspective. In becoming universal mimics that can imitate the behaviour of clearly rule driven processes, they become uninformative about the true nature of the phenomena they are "parroting" \cite{bender}. Enormous amounts of training data and advances in compute power have made the modern incarnation of artificial neural networks tremendously capable in solving certain problems that previously required human-like intelligence, but just like their predecessors, they have failed to deliver a revolution in our understanding of human cognition.

\bibliography{sn-bibliography}


\end{document}